\documentclass[10pt,twocolumn,letterpaper]{article}

\usepackage{cvpr}              

\usepackage{graphicx}
\usepackage{amsmath}
\usepackage{amssymb}
\usepackage{booktabs}
\usepackage{CJKutf8}

\usepackage[pagebackref,breaklinks,colorlinks]{hyperref}

\usepackage[capitalize]{cleveref}
\crefname{section}{Sec.}{Secs.}
\Crefname{section}{Section}{Sections}
\Crefname{table}{Table}{Tables}
\crefname{table}{Tab.}{Tabs.}

\def\cvprPaperID{*****} 
\def\confName{CVPR}
\def\confYear{2025}

\begin{document}
\begin{CJK}{UTF8}{gbsn}

\title{IMR: Iterative Mode-World Weighted Regression for Multi-Agent Trajectory Prediction}

\author{
Honglin Wang\\
EACON\\
Fujian, China\\
{\tt\small wanghonglin921@gmail.com}
\and
Shiyao Pan\\
EACON\\
Fujian, China\\
{\tt\small dcspsy@gmail.com}
\and
Yun-Fu Liu\\
EACON\\
Fujian, China\\
{\tt\small yunfuliu@gmail.com}
}
\maketitle

\begin{abstract}

   Multi-agent motion prediction is essential for automated vehicles to understand the intentions of surrounding vehicles. However, previous prediction-based and anchor-based methods have limitations in mode diversity and prediction accuracy, respectively. These limitations may cause inadequate safety assessments and behavioral deviations in automated vehicles. To address this issue, a mode-world weighted regression loss is proposed to bridge the gap between these features. Specifically, this approach mitigates mode collapse while simultaneously improving world ranking and top-1 confidence. Furthermore, the proposed iterative decoder improves prediction accuracy by recurrently and segmentally generating trajectories. Experimental results show the proposed method ranks first in the Argoverse 2 multi-agent motion forecasting benchmark against other methods.

\end{abstract}

\section{Introduction}
\label{sec:intro}

Motion prediction is crucial for autonomous driving technology, establishing the foundation for achieving high reliability and safety in autonomous vehicles \cite{huang2023review}, \cite{huang2022survey}. Through accurate prediction of future trajectories of road entities (\eg, vehicles, bicycles, pedestrians), it facilitates real-time environmental perception and dynamic analysis. This capability enhances path planning and obstacle avoidance precision, mitigates accident risks, and improves overall driving safety, contributing to smoother navigation.

Previous prediction-based methods (\eg, QCNet \cite{zhou2023query} and Forecast-MAE \cite{cheng2023forecast}) in multi-agent motion prediction are prone to mode collapse in complex scenarios. Anchor-based approaches (\eg, MTR \cite{shi2022motion} and TNT \cite{zhao2021tnt}) mitigate this issue at the expense of prediction accuracy. To resolve this trade-off, we propose a mode-world weighted regression loss within the prediction-based framework. Experiments demonstrate enhanced mode diversity without compromising accuracy, thereby bridging the performance gap across paradigms. Crucially, it mitigates mode collapse while concurrently improving the accuracy of world ranking and top-1 confidence.

Furthermore, we note that the previous state-of-the-art method QCNeXt \cite{zhou2023qcnext} on the Argoverse 2 \cite{wilson2023argoverse} multi-agent motion forecasting benchmark, employs a proposal-refinement decoding architecture. This method further refines proposed trajectories to generate more accurate predictions. However, when initial trajectories substantially deviate from ground truth, refinement layers struggle to effectively capture correct offsets due to excessive errors. Consequently, we propose an iterative decoder, and each iteration directly outputs trajectory coordinates rather than offsets. Simultaneously, encoded features, decoded outputs, and predicted trajectory from each iteration is propagated to next iteration, maximizing intermediate information utilization. This recurrent and segmented method facilitates iterative trajectory optimization.

\begin{figure*} 
  \centering
  \includegraphics[width=\textwidth]{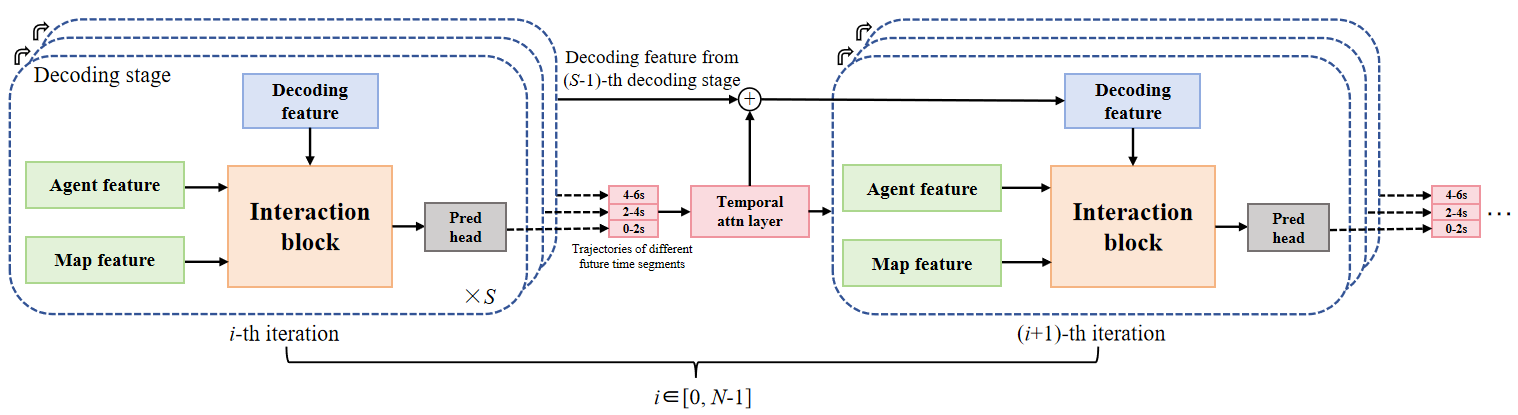} 
  \caption{
  The architecture of the proposed iterative decoder. The arrows shown around the top-left corner indicate the transmission of decoding features between different decoding stages.}
  \label{fig:myfigure}
\end{figure*}

\begin{figure}
  \centering
  \includegraphics[width=0.285\textwidth]{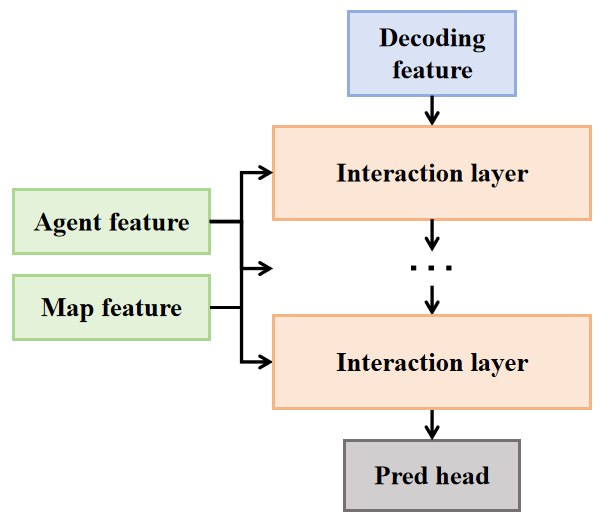} 
  \caption{
  Details of the interaction block shown in Fig. 1, in which the number of layers is denoted as $L$.}
  \label{fig:myfigure}
\end{figure}

\section{Approach}
\label{sec:formatting}

The proposed method is introduced in this section. First, the scene representation approach and encoder design are described. Second, the architecture and components of the iterative decoder are elaborated. Finally, the mode-world weighted regression loss is presented.

\subsection{Scene representation and encoder}

We treat the historical trajectory of each agent at every timestep as an independent element. Using the element as the origin, we construct a polar coordinate system and select agents within a preset radius as interaction objects. For each interaction pair, we compute 3D spatial relationships (relative distance, azimuth, orientation) to construct a dynamic attention weighted matrix.

The graph attention network-based encoder models each element and their interactions as a graph structure. Each element is represented as a node, with inter-node connections denoting interaction weights. By modeling multi-source interactions, this captures implicit relationships between different elements.

\subsection{Iterative decoder}

Figure 1 illustrates the architecture of the iterative decoder. The agent and map features represent the outputs from the encoder. The iterative decoder executes $N$ iterations. Each iteration comprises $S$ decoding stages designed to progressively decode trajectory segments, which are then concatenated into a full trajectory. The predicted trajectory from each iteration is processed by a temporal attention layer and then combined with the decoding feature to serve as input for the subsequent iteration. 

As depicted in Fig. 2, each interaction block comprises $L$ stacked interaction layers and it outputs predicted trajectory through a prediction head as formulated below, 

\begin{equation}
  F_{i} \in R^{K \times D}, \: i = 0
  \label{eq:important}
\end{equation}

\begin{equation}
  F_{i} = F^{'}_{i - 1} + LSTM(y_{i - 1}), \: i > 0
  \label{eq:important}
\end{equation}
where $F_{i}$ and $F^{'}_{i}$ denote the input and output features of the $i$-th iteration, respectively. $K$ is the number of predicted world, $D$ is the hidden size, $LSTM$ indicates the LSTM layer, and $y_{i}$ represents the predicted trajectory of the $i$-th iteration.

In each iteration, predicted trajectory is executed in segments. After the prediction of each segment of the trajectory is completed each time, the segments are connected to generate the complete trajectory. The equation shows below, 

\begin{equation}
  F^{'j}_{i} = GAT(F_{i}, A, M), \: j = 0
  \label{eq:important}
\end{equation}

\begin{equation}
  F^{'j}_{i} = GAT(F^{'j - 1}_{i}, A, M), \: j > 0
  \label{eq:important}
\end{equation}
where $GAT$ denotes the hierarchical multivariate Graph Attention Networks layer, $A$ and $M$ represent the encoding feature of agents and maps, respectively. $F^{'j}_{i}$ denotes the decoding feature for the $j$-th decoding stage in the $i$-th iteration.

Segmented processing can break down the complex trajectory prediction task into multiple simple sub-tasks, making the prediction process more controllable. Meanwhile, each prediction segment is known with the result of the previous segment, forming a dynamic interaction relationship. The iterative process model continuously explores the potential patterns in the data by constantly reusing the complete trajectory generated last time, and further optimizes the prediction results.

It is worth noting that we set the output of each decoding stage to the position coordinate value of the predicted trajectory instead of the offset relative to the previous decoding stage. This helps to avoid the uncorrectable consequences of the incorrect prediction of the previous decoding stage on the current decoding stage.

\begin{table*}
  \centering
  \begin{tabular}{@{}lccccccc@{}}
    \toprule
    Method &
    $\text{avgMinFDE}_{\text{6}}$ &
    $\text{avgMinFDE}_{\text{1}}$ &
    $\text{actorMR}_{\text{6}}$ &
    $\text{avgMinADE}_{\text{6}}$ &
    $\text{avgMinADE}_{\text{1}}$ &
    $\text{avgBrierMinFDE}_{\text{6}}$ &
    $\text{actorCR}_{\text{6}}$ \\
    \midrule
    LRP & 1.21 & 2.57 & 0.16 & 0.58 & 1.05 & 1.85 & 0.01 \\
    DONUT & 1.17 & 2.73 & 0.15 & 0.58 & 1.13 & 1.83 & \textbf{0.01} \\
    SEPT \cite{lan2023sept} & 1.14 & 3.22 & 0.15 & 0.55 & 1.27 & 1.80 & 0.01 \\
    LOF \cite{wang2024futurenet}  & 1.25 & 2.34 & 0.18 & 0.58 & 0.96 & 1.68 & 0.02 \\
    QCNeXt \cite{zhou2023qcnext} & \textbf{1.02} & \textbf{2.29} & \textbf{0.13} & \textbf{0.50} & 0.94 & 1.65 & 0.01 \\
    IMR (ours) & 1.08 & 2.29 & 0.15 & 0.51 & \textbf{0.92} & \textbf{1.59} & 0.01 \\
    \bottomrule
  \end{tabular}
  \caption{Quantitative results on the Argoverse 2 multi-agent motion forecasting benchmark.}
  \label{tab:example}
\end{table*}

\begin{table*}
  \centering
  \begin{tabular}{@{}lccccccc@{}}
    \toprule
    Method &
    $\text{MinFDE}_{\text{6}}$ &
    $\text{MinFDE}_{\text{1}}$ &
    $\text{MinADE}_{\text{6}}$ &
    $\text{MinADE}_{\text{1}}$ &
    $\text{MR}_{\text{6}}$ &
    $\text{MR}_{\text{1}}$ &
    $\text{brier-MinFDE}_{\text{6}}$ \\
    \midrule
    DeMo \cite{zhang2024decoupling} & 1.11 & 3.70 & 0.60 & 1.49 & 0.12 & 0.55 & 1.73 \\
    Polaris & 1.11 & 3.70 & 0.61 & 1.53 & 0.12 & 0.54 & 1.71 \\
    NDPNet & 1.09 & 3.71 & 0.58 & 1.47 & 0.13 & 0.53 & 1.71 \\
    SEPT++ \cite{lan2023sept}  & 1.09 & \textbf{3.51} & 0.57 & \textbf{1.41} & 0.13 & 0.51 & 1.65 \\
    LOF \cite{wang2024futurenet} & 1.07 & 3.63 & 0.58 & 1.46 & \textbf{0.12} & 0.51 & 1.63 \\
    IMR (ours) & \textbf{1.08} & 3.54 & \textbf{0.57} & 1.42 & 0.12 & \textbf{0.51} & \textbf{1.63} \\
    \bottomrule
  \end{tabular}
  \caption{Quantitative results on the Argoverse 2 single agent motion forecasting benchmark.}
  \label{tab:example}
\end{table*}

\begin{figure*}
  \centering
  \begin{subfigure}{0.3\linewidth}
    \includegraphics[width=\textwidth]{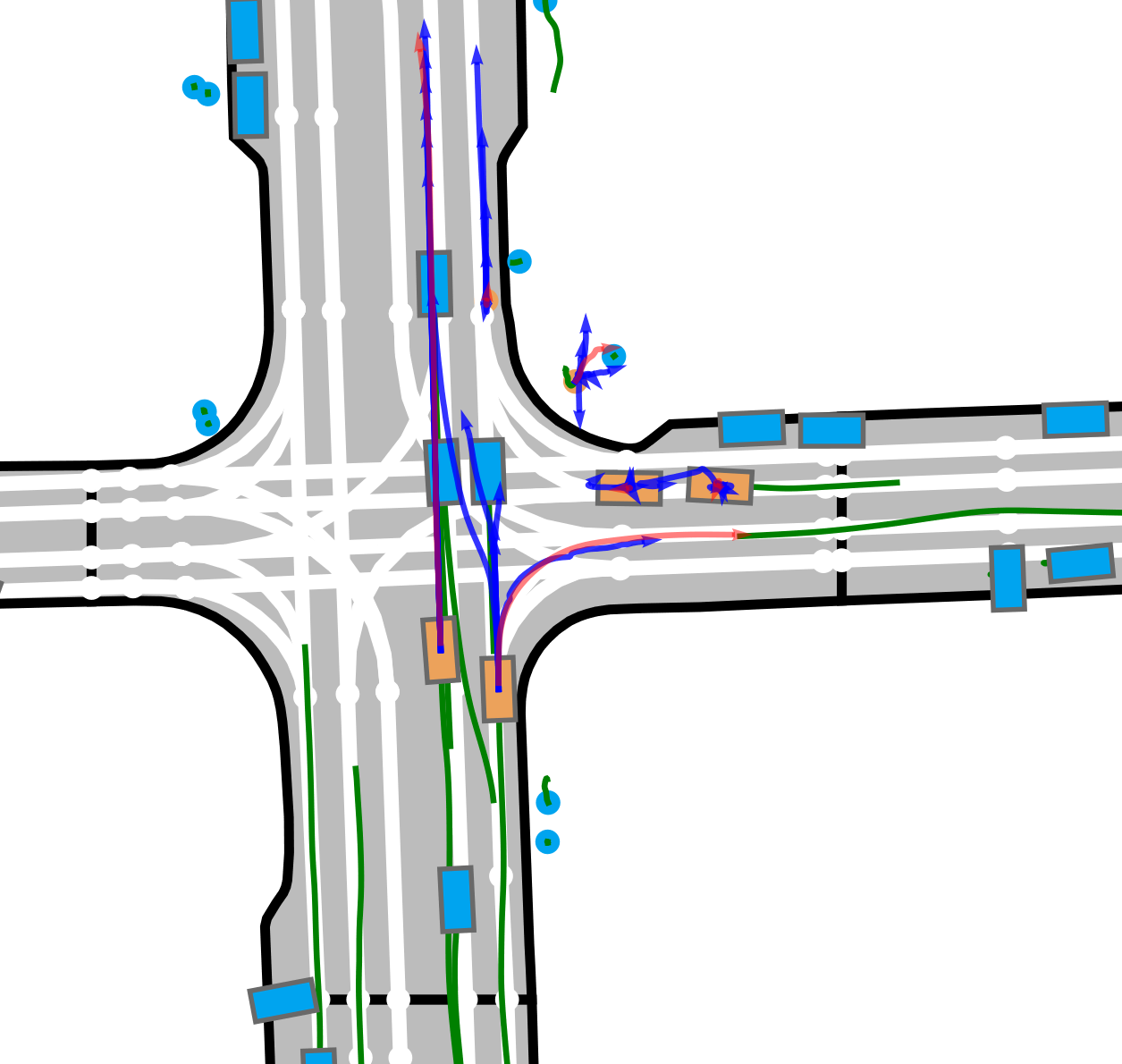}
    \label{fig:short-a}
    \caption{Case 1: mode-wise loss}
  \end{subfigure}
  \begin{subfigure}{0.3\linewidth}
    \includegraphics[width=\textwidth]{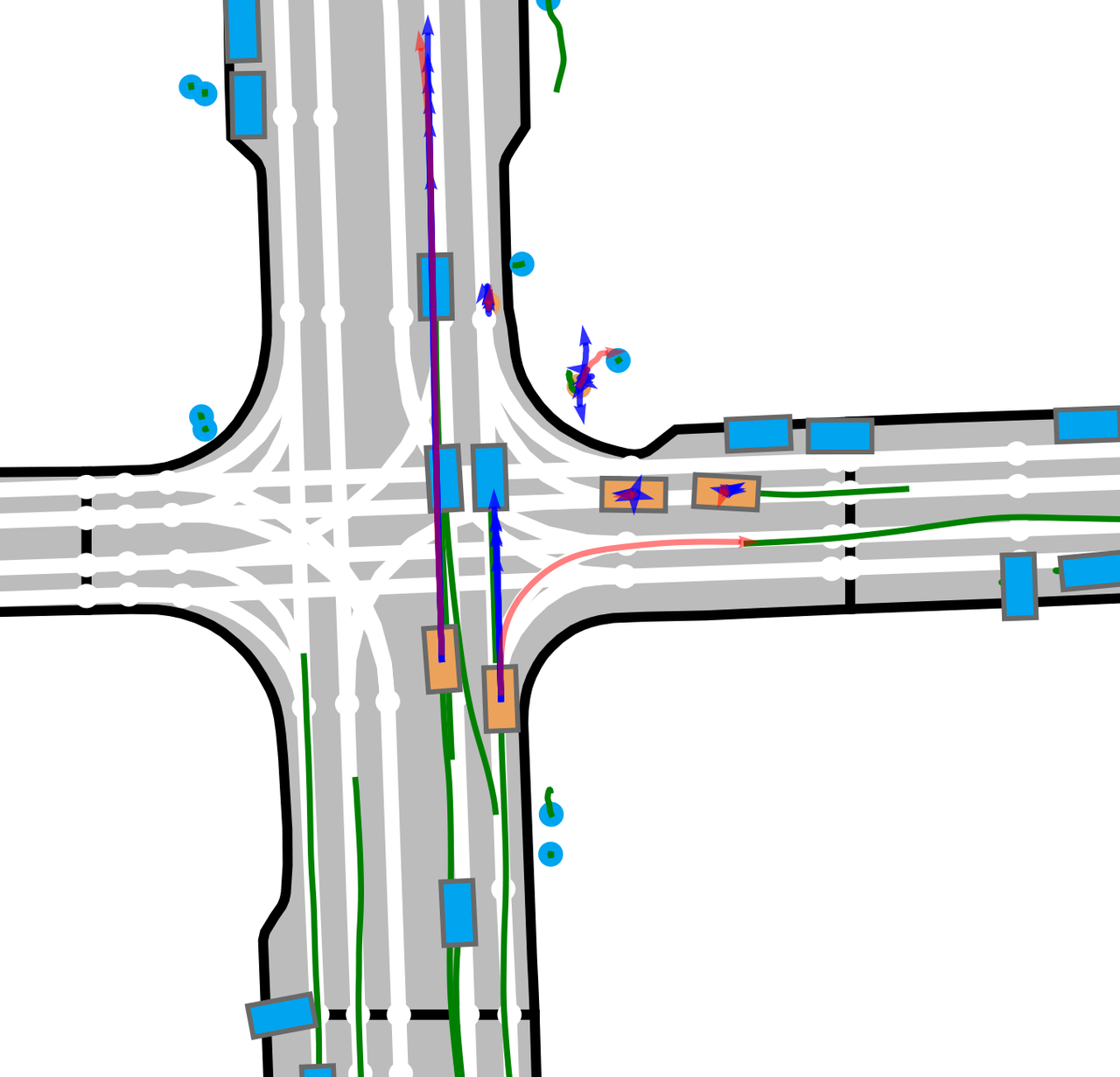}
    \label{fig:short-a}
    \caption{Case 1: world-wise loss}
  \end{subfigure}
  \begin{subfigure}{0.3\linewidth}
    \includegraphics[width=\textwidth]{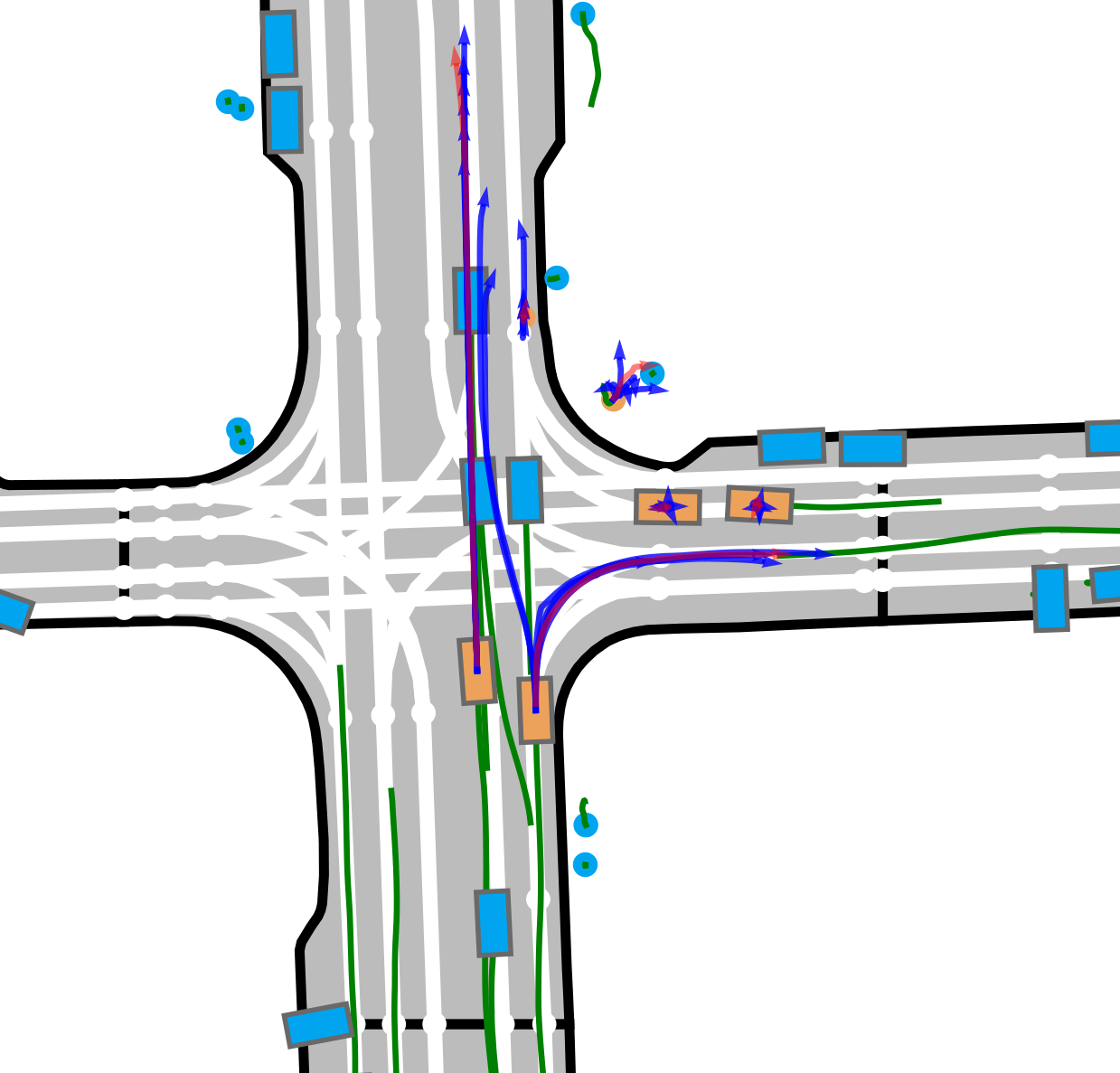}
    \label{fig:short-a}
    \caption{Case 1: mode-world loss}
  \end{subfigure}
  \begin{subfigure}{0.3\linewidth}
    \includegraphics[width=\textwidth]{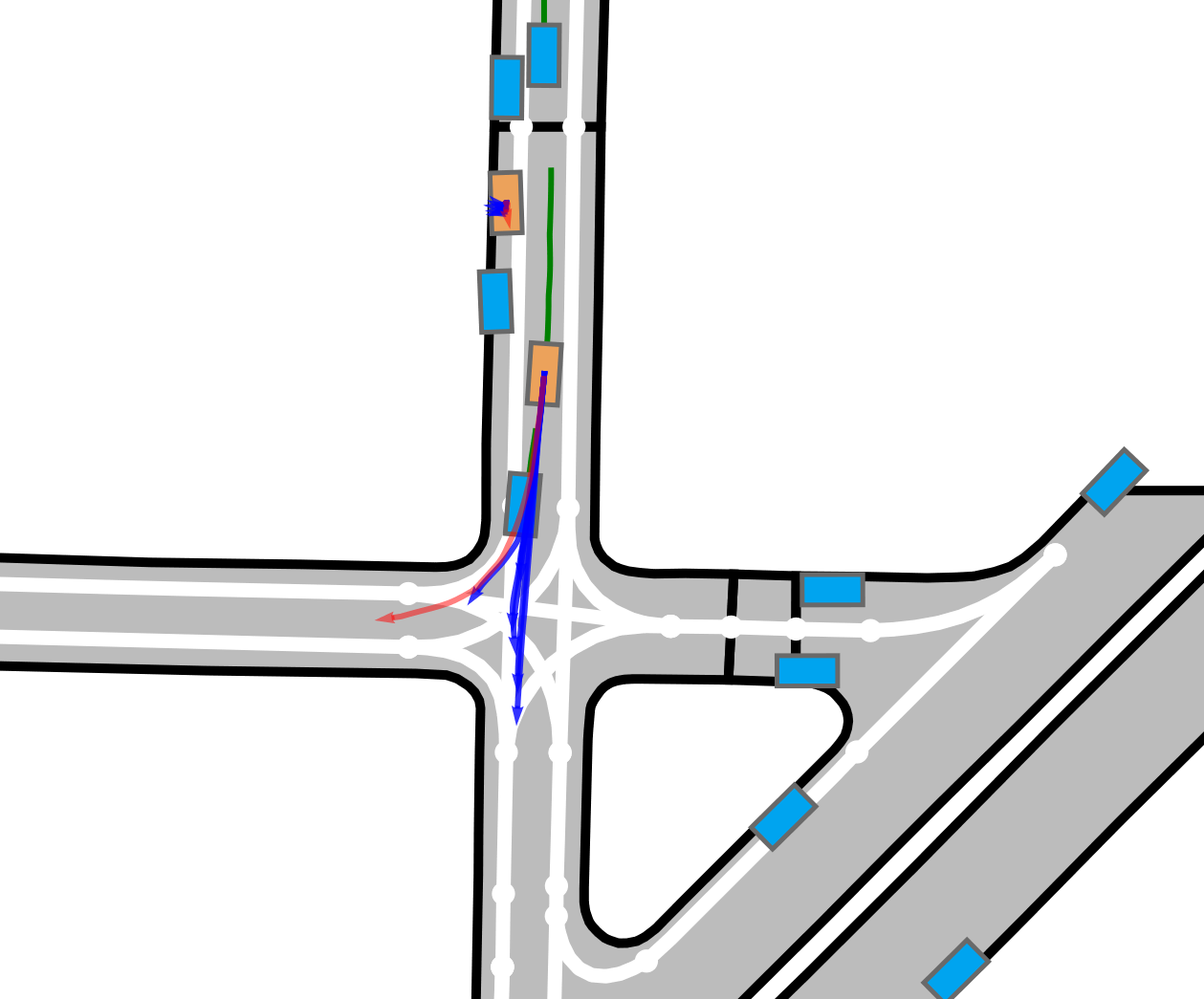}
    \label{fig:short-a}
    \caption{Case 2: mode-wise loss}
  \end{subfigure}
  \begin{subfigure}{0.3\linewidth}
    \includegraphics[width=\textwidth]{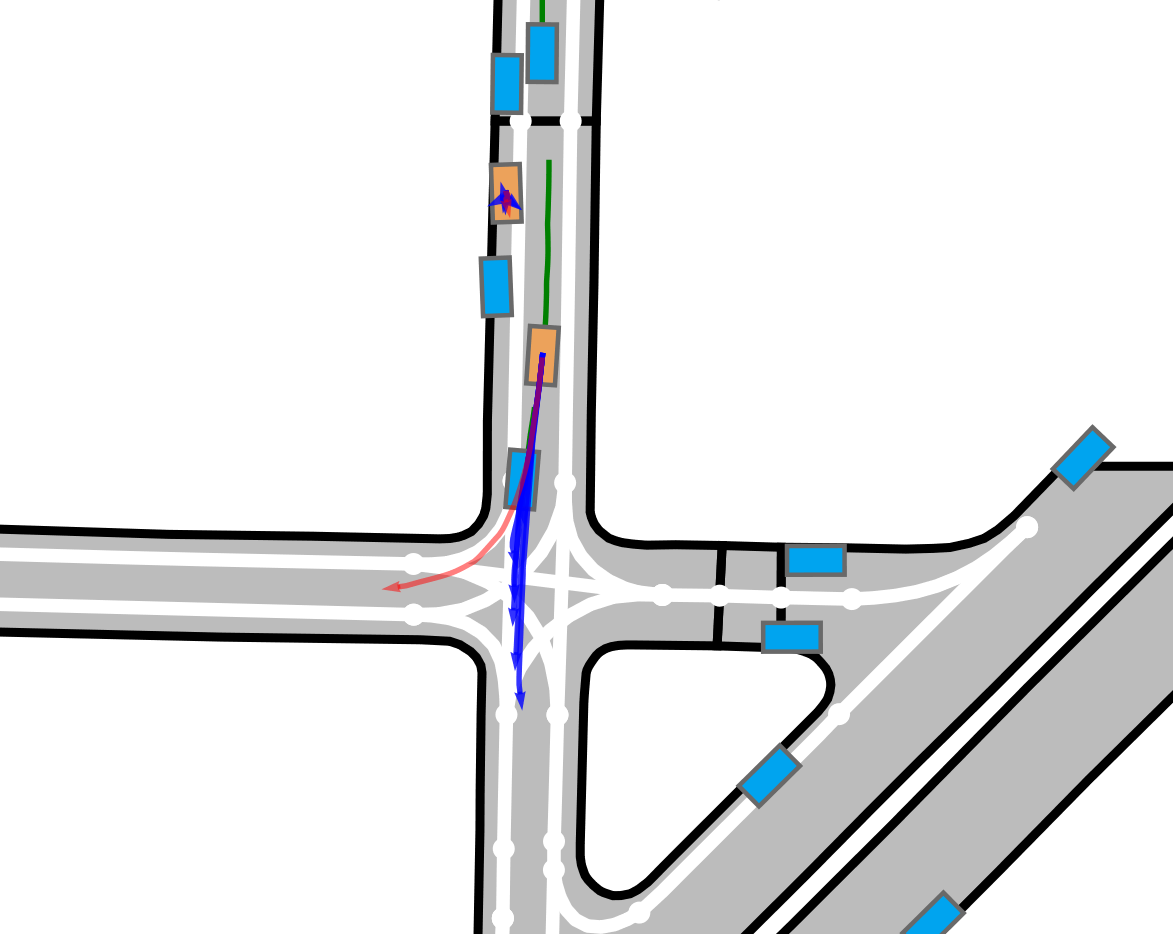}
    \label{fig:short-a}
    \caption{Case 2: world-wise loss}
  \end{subfigure}
  \begin{subfigure}{0.3\linewidth}
    \includegraphics[width=\textwidth]{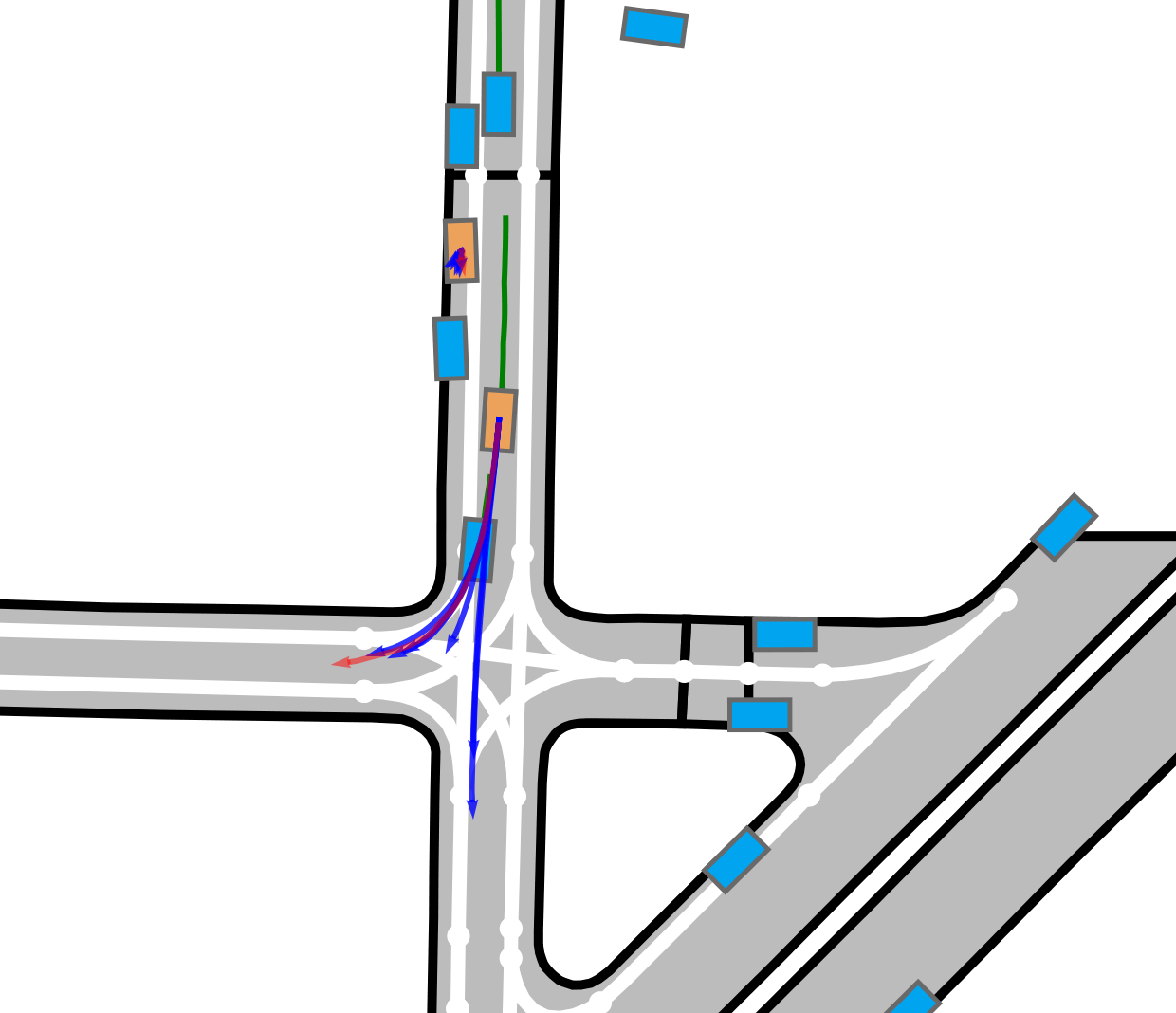}
    \label{fig:short-a}
    \caption{Case 2: mode-world loss}
  \end{subfigure}
  \caption{Visualization comparison of different types of regression loss on Argoverse 2 validation set. Each row represents a different case, and the results of three types of loss involving mode-wise, world-wise, and the proposed mode-world regression loss are shown in different columns.}
\end{figure*}

\subsection{Mode-world weighted regression loss}

In order to generate the world-wise confidence scores to adapt to the joint trajectory prediction task, each world should contain the joint trajectories of all agents as well as a corresponding confidence score to this world. During the training process, we use laplace loss and focal loss as the regression loss function and classification loss function respectively.  Among them, the final regression loss value is obtained by the weighted method. The mode-wise regression loss and the world-wise regression loss are calculated together and weighted. The equation is formulated as follows,

\begin{equation}
  L_{reg} = \omega_{mode} \cdot L^{mode}_{reg} + \omega_{world} \cdot L^{world}_{reg}
  \label{eq:important}
\end{equation}
where $L_{reg}$ represents the final regression loss value, $\omega_{mode}$ and $\omega_{world}$ respectively represent the mode-wise and world-wise regression loss weights, $L^{mode}_{reg}$ and $L^{world}_{reg}$ represent the mode-wise regression loss function value and the world-wise regression loss value.

For mode regression, a mode winner-takes-all strategy is adopted to minimize losses. The winner is defined as the best mode-wise trajectory in each agent. Among them, the selection method of the optimal mode-wise trajectory is obtained by weighting the minimum maximum displacement error and the minimum average displacement error, as shown in the following equation:

\begin{equation}
  m = \omega_{ADE} \cdot ADE + \omega_{MDE} \cdot MDE
  \label{eq:important}
\end{equation}
where $m$ represents the optimal mode-wise trajectory. $\omega_{ADE}$ and $\omega_{MDE}$ represent the weights corresponding to the two kinds of errors, and $ADE$ and $MDE$ are the average and maximum displacement error, respectively.

\section{Experiments}
\label{sec:intro}

\subsection{Implementation details}

The length of the predicted trajectory is supposed to 6s as defined in Argoverse 2. The hidden size $D$ is set to 128, and the number of iterations $N$ is set to 6. For each iteration, the predicted trajectory is set to 3 segments at equal time intervals in a chronological order, and the number of stacked interaction layers $L$ is set to 3 as well. Notably, the initial decoding feature is randomly initialized for a cold start. The number of predicted worlds $K$ is 6. $\omega_{mode}$ and $\omega_{world}$ are 0.5 and 1, respectively. We train the model using AdamW for 60 epochs on the Argoverse 2 motion forecasting dataset with batch size 32. The initial learning rate is $5 \times 10^{-4}$, with weight decay and dropout both at 0.1. Through cosine annealing, the learning rate decays to 0.

\subsection{Ensemble}

We train 10 models with different random seeds, producing a total of 60 predicted worlds. These world-wise predictions are then integrated using the weighted k-means algorithm introduced in QCNeXt \cite{zhou2023qcnext}.

\subsection{Quantitative results}

We compare our approach to other methods on the Argoverse 2 multi-agent motion forecasting benchmark. As shown in Table 1, our method achieves a 0.06 improvement in $avgBrierMinFDE_6$ over the previous SOTA QCNeXt. Furthermore, we also verified the effectiveness of the proposed method in the Argoverse 2 single agent motion forecasting benchmark as listed in Table 2. Our model achieves a competitive performance in contrast to LOF \cite{wang2024futurenet} as well. 

\subsection{Qualitative results}

Figure 3 presents a visual comparison of different types of regression losses. In these figures, green lines depict historical trajectories of orange-colored interested agents. Orange and blue lines indicate ground truth and the predicted future trajectories, respectively. As shown in the mode-wise loss results of Figs. 3(a) and (d), the predicted trajectories exhibit high diversity yet a lower accuracy is introduced due to the length inconsistency with the ground truth. Figures 3(b) and (e) present the results of the world-wise regression and they reveal significant mode collapse as represented by absence of predictions matching ground truth. In contrast, the proposed mode-world weighted regression loss mitigates mode collapse while improving prediction accuracy as shown in Figs. 3(c) and (f).

\section{Conclusions}
\label{sec:intro}

In this work, a multi-agent motion prediction method is proposed to address insufficient mode diversity and low prediction accuracy. Experimental results demonstrate that the weighted regression loss mitigates mode collapse and resolves the diversity-accuracy trade-off. Furthermore, it enhances world ranking accuracy and top-1 confidence. By generating trajectories through a recurrent and segmented method, the decoder enables the model to learn from previous information, gradually improving the prediction accuracy. However, this approach incurs high computational complexity in decoding, limiting real-time applicability. As demonstrated in experimental results, our method ranks first on the Argoverse 2 multi-agent motion forecasting benchmark and achieves first-tier performance on the single-agent benchmark.

{\small
\bibliographystyle{ieee_fullname}
\bibliography{egbib}
}
\end{CJK}
\end{document}